\documentclass[conference]{IEEEtran}
\pdfoutput=1

\usepackage{amsmath,graphicx}
\usepackage{cite}
\usepackage{amssymb,amsfonts}
\usepackage{algorithmic}
\usepackage{textcomp}
\usepackage{xcolor} 
\usepackage{booktabs}
\usepackage{makecell}
\usepackage{multirow}
\usepackage{capt-of} 
\usepackage[pdftitle={Neural Codec Wrapping}]{hyperref}

\newcommand{\eps}{\epsilon}

\title{A Projection-Based Surrogate Gradient Interpretation for Neural Codec Wrappers}

\author{
\IEEEauthorblockN{
Esteban Pesnel\IEEEauthorrefmark{1}\IEEEauthorrefmark{2}, 
Julien Le Tanou\IEEEauthorrefmark{1}, 
Michael Ropert\IEEEauthorrefmark{1},
Aline Roumy\IEEEauthorrefmark{2}, 
Thomas Maugey\IEEEauthorrefmark{2}
}
\IEEEauthorblockA{
\IEEEauthorrefmark{1}MediaKind, Rennes, France \\
\{esteban.pesnel, julien.letanou, michael.ropert\}@mediakind.com
}
\IEEEauthorblockA{
\IEEEauthorrefmark{2}INRIA, Rennes, France \\
\{aline.roumy, thomas.maugey\}@inria.fr
}
}

\begin{document}
\maketitle

\begin{abstract}
Neural wrappers are learned pre- and post-processing networks designed to enhance the performance of conventional video codecs. Although these approaches can significantly improve compression efficiency, training them remains challenging due to the non-differentiability of video codecs, which arises from the multiple discrete decisions involved in the encoding process. Surrogate gradients have recently emerged as an effective solution for enabling end-to-end learning with conventional codecs. They offer two main advantages: they avoid training an additional network to mimic the codec, and they can improve compression performance. In particular, the recently proposed SCALED method, which leverages the true compression error, has shown strong results for training neural pre-processors such as downscalers. However, this SCALED gradient was originally introduced as a reparameterization trick, which limits its interpretability. In this paper, we show that this surrogate gradient can be interpreted as a first-order local approximation of the video codec, providing insight into its effectiveness. We further demonstrate that it is effective not only for learning downscaling operations, but also for the more challenging task of full neural wrapping with pre- and post-processing networks. Finally, we show that the approach generalizes well across different video codecs, quality factors, and tasks, including multiple downscaling ratios, yielding BD-Rate (PSNR) reductions of up to -23.59\% on x264 and -20.07\% on VVenC relative to standard resampling baselines.
\end{abstract}
\section{Introduction}

Neural network-based wrappers have emerged as effective pre- and post-processing tools for augmenting conventional video codecs, attracting significant interest in recent years. For instance, version 3 of the Versatile Supplemental Enhancement Information Messages for coded video bitstreams (VSEI), Rec. ITU-T H.274 \cite{ITUH274}, standardizes the signaling of Neural-Network Post-Filtering (NNPF) information. However, end-to-end (E2E) training of such neural wrappers remains challenging because conventional codecs prevent reliable gradient computation during backpropagation.

This difficulty arises from the non-differentiability of several processing blocks in conventional codecs. Quantization operations, for instance, have gradients that are zero almost everywhere, preventing the backpropagated error from reaching the preprocessing filter, an issue already addressed for E2E neural codecs via additive noise approximations~\cite{DCVC,DVC}. Other coding decisions, such as block partitioning, intra/inter prediction mode selection, and motion vector estimation, are inherently discrete and introduce additional non-differentiable steps.
To overcome this, several works replace the codec during training with a differentiable virtual codec, typically a neural network trained to mimic the true codec~\cite{ACN, ByteDance, tian2021self, CNN-RD, Shangai, zhao2019learning, qiu2021codec, AIDN, yang2023self}.  
However, the resulting training process becomes dependent on the target codec and quality parameter, requiring multiple trainings and the storage of several models.

To address the issue of multiple trained models, another line of research has investigated differentiable yet non-learned proxies for approximating video codec behavior. 
A first category of approaches approximates the codec by the identity function in order to train the downscaler \cite{CAR, CNN-CR}. A second category \cite{Google2024} models the codec using  conventional codec operations such as linear prediction, DCT, and quantization while avoiding non-differentiability by fixing coding decisions, e.g., fixed block sizes and fixed intra/inter prediction modes. 
More recently, in \cite{pesnel2025scaled}, we proposed an alternative strategy that also avoids the learning of a virtual codec, and exploits, during training, realizations of the coding noise generated by the true codec to improve its fidelity. This approach has demonstrated strong reliability in practice, in the context of learning a prefilter (downscaler), where it outperforms state-of-the-art methods.

Despite its strong empirical performance, the surrogate gradient \cite{pesnel2025scaled} lacks interpretability. It is introduced as a reparameterization trick \cite{Mack} in which the gradient of the distortion term is replaced by the gradient of the standard deviation of the coding error. In this paper, we provide a novel interpretation of this surrogate gradient. We show that it can be derived as the gradient of a first-order approximation of the true codec. Furthermore, we demonstrate that this surrogate gradient is effective not only for learning downscalers, as previously shown in \cite{pesnel2025scaled}, but also for the more challenging task of training neural wrappers. 

\begin{figure*}[ht]
  \centering
  \includegraphics[width=0.95\textwidth]{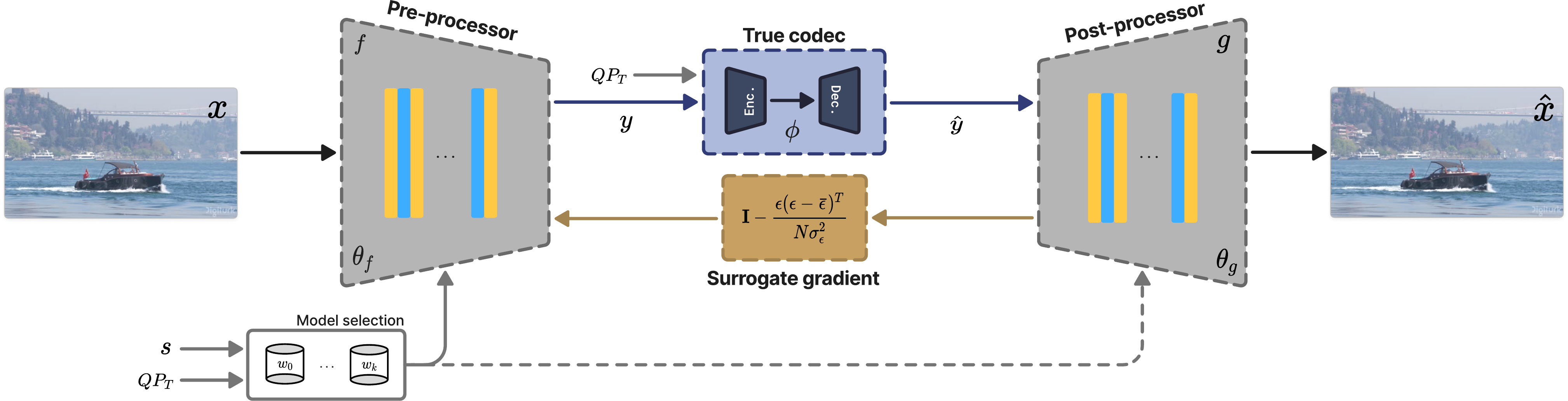}
  \caption{Overview of the proposed neural wrapper training pipeline. Blue arrows denote the forward pass ($\boldsymbol{x} \to \boldsymbol{\hat{x}}$), while yellow arrows indicate the backward surrogate gradient approximation~\cite{pesnel2025scaled}.}
    \label{fig:wide-image}
\end{figure*}

\section{Gradient approximation for neural wrapper optimization}

\subsection{End-to-end learning}

The inference scheme of neural wrappers involves sandwiching a standard video codec between neural pre- and post-processors. 
Let $\boldsymbol{x} \in \mathbb{R}^{N}$ be the input content. The pre-processor $f$ transforms $\boldsymbol{x}$ into a latent representation $\boldsymbol{y} = f(\boldsymbol{x};\boldsymbol{\theta}_f)$. This representation is compressed by a standard, non-differentiable video codec $\phi$ (e.g., H.264/AVC), producing the decoded latent $\boldsymbol{\hat{y}} = \phi(\boldsymbol{y})$. Finally, the post-processor $g$ reconstructs the output video $\boldsymbol{\hat{x}} = g(\boldsymbol{\hat{y}};\boldsymbol{\theta}_g)$.
The optimization goal is to minimize a joint rate-distortion objective:

\begin{equation}
    \boldsymbol{\theta}_f^*,\boldsymbol{\theta}_g^*
    = \arg\min_{\boldsymbol{\theta}_f, \boldsymbol{\theta}_g} 
    D \left( \boldsymbol{x}, \boldsymbol{\hat{x}}
    \right)
    +\lambda R
    \label{Global objective}
\end{equation}
where $D$ denotes any distortion function, and $R$ is the encoding rate. 
However, direct end-to-end minimization of \eqref{Global objective} with gradient descent is intractable. Many codec operations such as quantization or mode decision are non-differentiable, breaking the gradient flow 
and preventing the back-propagation of the reconstruction error to the pre-processor.

%
%

\subsection{SCALED: Codec surrogate gradient using true compression error statistics}

To address the non-differentiability of the codec, the SCALED method proposed in \cite{pesnel2025scaled} relies on a reparameterization trick in which the codec error is replaced, during backpropagation, by its standard deviation. More precisely, let $\boldsymbol{y} \in \mathbb{R}^N$ be the input to the codec and let $\boldsymbol{\hat{y}}$ be the decoded signal such that 
\begin{equation}
    \boldsymbol{\hat{y}} = \phi (\boldsymbol{{y}}) = \boldsymbol{y} + \boldsymbol{\epsilon},
\label{eq:def_codec}
\end{equation}
where $\boldsymbol{\epsilon} \in \mathbb{R}^N$ represents the compression error. During training, the SCALED approach replaces the codec $\boldsymbol{\hat{y}} = \phi(\boldsymbol{y})$ with:
\begin{equation}
    \hat \phi (\boldsymbol{{y}}) = \boldsymbol{y} + \operatorname{sg}(\boldsymbol{\epsilon})\frac{\sigma_{\boldsymbol{\epsilon}}}{\operatorname{sg}(\sigma_{\boldsymbol{\epsilon}})},
\label{SCALED_fwd}
\end{equation}
\noindent where $\operatorname{sg}(\cdot)$ denotes the stop-gradient operator, which acts as the identity function in the forward pass and returns zero in the backward pass. 
%
Therefore, \eqref{SCALED_fwd} can be rewritten as
\begin{subequations}
\begin{align}
\text{Forward}: & \ \boldsymbol{y} + \boldsymbol{\epsilon} =\boldsymbol{\hat{y}} 
= \phi(\boldsymbol{y})   \\
\text{Backward}: & \ 
\dfrac{\partial\,\hat \phi(\boldsymbol{y})}{\partial \boldsymbol{y}} = \mathbf{I} - \frac{\boldsymbol{\epsilon} (\boldsymbol{\epsilon} -\bar{\boldsymbol{\epsilon}})^T}{N\sigma^2_{\boldsymbol{\epsilon}}}  = \mathbf{J}_{\text{SCALED}}  
\end{align}
\label{eq:sgcodec}
\end{subequations}
where $\bar{\boldsymbol{\epsilon}} \in \mathbb{R}^N$ is the constant mean vector of $\boldsymbol{\eps}$, and $\sigma_{\eps}$ is the empirical standard deviation:
\begin{equation}
   \sigma_{\eps}^2= \frac{1}{N} (\boldsymbol{\eps}-\bar{\boldsymbol{\epsilon}})^T\boldsymbol{\eps}.
    \label{empirical_std}
\end{equation}
The derivation of the Jacobian $\mathbf{J}_{\text{SCALED}}$ can be found in \cite{pesnel2025scaled}.

The training of a neural wrapper is performed through gradient backpropagation, and \eqref{eq:sgcodec} details how this backpropagation is carried out through the codec. This learning procedure was shown to be effective and produce preprocessors with better compression performance than other methods, such as \cite{Google2024}. This can be explained by the fact that the forward pass (\ref{eq:sgcodec}a) of $\hat \phi$ corresponds exactly to the true codec, and that the backward pass (\ref{eq:sgcodec}b) depends on the noise realization of the codec. However, there is no evidence regarding the accuracy of the model in the backward pass, which is only assessed numerically through the coding performance of the preprocessor. Therefore, in the next section, we propose a model directly inspired by the properties of the codec in order to improve interpretability. Finally, we show that this new model is equivalent to the SCALED model derived from a parametrization trick.

\section{A Projection-based Gradient approximation and Geometric interpretation}
\label{sec:demo}

The goal of this section is to model a codec in such a way that it is locally reliable and that the gradient is easy to compute with respect to the output of the true codec. To this end, we propose several properties of codecs and numerically test the validity of these hypotheses. We then propose a linear model that satisfies these properties and derive its parameters.

\subsection{Codec properties}
\label{sec:codecProp}

\begin{figure}
    \centering
    \includegraphics[width=1.0\linewidth]{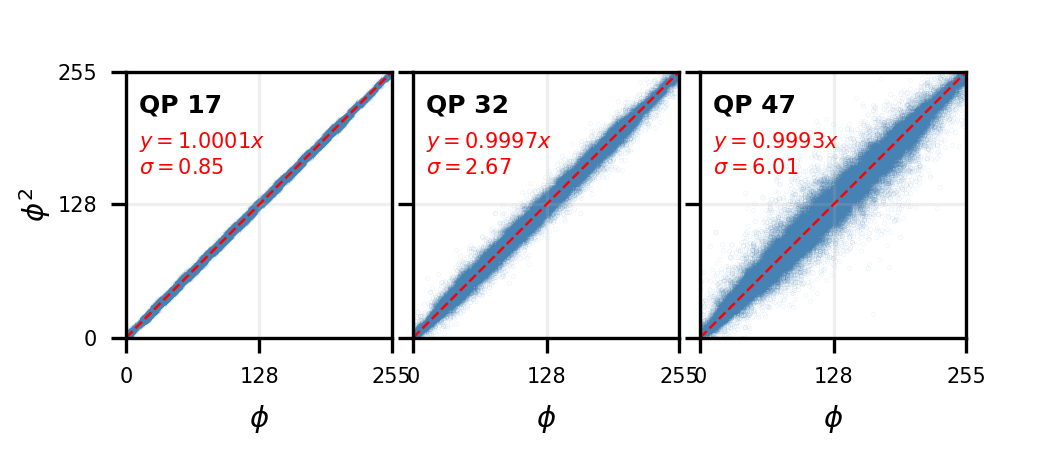}
    \vspace{-0.8cm}
    \caption{Codec idempotence: $\phi^2$ vs $\phi$ (x264, \textit{medium} preset, average over 30 sequences of Google dataset \cite{Google2024}). Near-unity slope confirms $\phi^2 \approx \phi$. $\sigma$ is the residual std from the fitted line $y = ax$.}
    \label{fig:idempotence}
\end{figure}

\textbf{Idempotence:} The processing that introduces errors in a video codec is quantization, which is idempotent. A complete codec, however, is not strictly idempotent. Indeed, a decoded image may have a different partitioning from that of the original image and will therefore undergo different processing. Here, we investigate whether a codec is nevertheless close to being idempotent. Figure~\ref{fig:idempotence} shows the output of the codec applied twice ($\phi^2$) as a function of the output of the codec applied once ($\phi$). The identity line corresponds to the case of strict idempotence. Interestingly, we observe that, to a first approximation, the codec is close to being idempotent, especially at low QP.

\textbf{Intensity-shift invariance:} 
The invariance of entropy and differential entropy to a shift \cite{CoverThomas06}, together with the invariance of the distortion when both the input and output are shifted, suggests that a codec might be invariant to shifts in pixel intensity values. We tested this hypothesis, and Table~\ref{tab:translation_invariance} shows that the codec is indeed close to being intensity-shift invariant.

\begin{figure}
    \centering
    \includegraphics[width=1\linewidth]{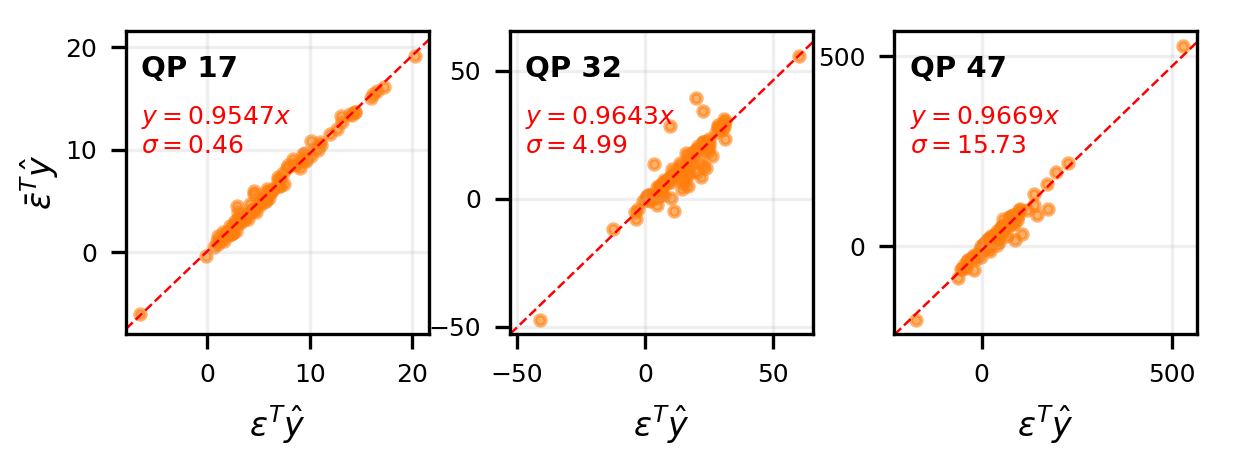}
    \vspace{-0.5cm}
    \caption{Orthogonality: $\boldsymbol{\eps}^T\boldsymbol{\hat{y}}$ vs $\bar{\boldsymbol{\epsilon}}^T\boldsymbol{\hat{y}}$ (x264, \textit{medium} preset, 100 sequences). Near-unity slope confirms $(\boldsymbol{\epsilon} -\bar{\boldsymbol{\epsilon}})^T\boldsymbol{\hat{y}} \approx 0$. Non-centered mean: points fall on the diagonal $\bar{\boldsymbol{\epsilon}}^T\boldsymbol{\hat{y}} \approx \boldsymbol{\epsilon}^T\boldsymbol{\hat{y}} \neq 0$ rather than on the axis, so $\bar{\boldsymbol{\epsilon}} \neq 0$ and $\boldsymbol{\epsilon}$ is not centered. $\sigma$ is the residual std from the fitted line $y = ax$.}
    \label{fig:eps_vs_eps_bar}
    \vspace{-0.4cm}
\end{figure}

\textbf{Non-centered compression error:} Compression errors are classically assumed to have zero mean. However, we measured the mean compression error for real codecs and observed that this assumption does not hold, as depicted in Figure \ref{fig:eps_vs_eps_bar}. A small but nonzero mean value exists, and this is of utmost importance to consider when back-propagating gradients.

\textbf{Orthogonality between centered codec error and codec output:} Orthogonality between codec error and codec output is a classical assumption. It is a key step in deriving the Shannon lower bound \cite{CoverThomas06}, which provides an approximation of the rate-distortion function, that is asymptotically tight at high rates \cite{Koch16, linder1994asymptotic}. Following the previous observation, we test the orthogonality between the centered codec error and the codec output. More precisely, we plot $\boldsymbol{\eps}^T\boldsymbol{\hat{y}}$ versus $\bar{\boldsymbol{\epsilon}}^T\boldsymbol{\hat{y}}$, where equality corresponds to orthogonality. Figure~\ref{fig:eps_vs_eps_bar} shows that orthogonality holds to a very good approximation.

\subsection{Modeling the codec as a linear projection}

We propose a first-order approximation of a codec. We therefore consider a linear operator. Moreover, due to the idempotence property of the codec (see Section~\ref{sec:codecProp}), we model the codec as a linear projection. Such a projection is completely characterized by: (i) a projection direction $\boldsymbol{u}$, and (ii)
the subspace onto which the signal is projected, characterized by a vector $\boldsymbol{v}$. More formally, a projector $\mathbf{P}$ satisfies $\forall \boldsymbol{y}$ and for some $\beta$:
\begin{subequations}\label{eq:proj}
\begin{align}
\mathbf{P}\boldsymbol{y} &= \boldsymbol{y} - \beta \boldsymbol{u}, \label{eq:proj1} \\
    \boldsymbol{v}^T \mathbf{P}\boldsymbol{y} &= 0,\label{eq:proj2}
\end{align}
\end{subequations}
which leads to 
\begin{equation}
    \hat \phi(\boldsymbol{y}) = \mathbf{P} \boldsymbol{y} 
    = \left( \mathbf{I} - \frac{\boldsymbol{u} \boldsymbol{v}^T}{\boldsymbol{v}^T \boldsymbol{u}} \right) \boldsymbol{y}. \label{eq:projector}
\end{equation}

\textbf{Determination of $\boldsymbol{u}$:} Since the codec output corresponds to a shift of the input by the error vector $\boldsymbol{\eps}$, by design $\boldsymbol{u}=\boldsymbol{\epsilon}$.\\

\textbf{Determination of $\boldsymbol{v}$:} First, we show that the intensity-shift invariance hypothesis (see Section~\ref{sec:codecProp}) implies that the vector $\boldsymbol{v}$ is centered. We assume that, like the codec $\phi(\boldsymbol{y})$, its model $\hat \phi(\boldsymbol{y})=\mathbf{P}\boldsymbol{y}$ is also intensity-shift invariant. Therefore, 
\begin{align}
    \mathbf{P}(\boldsymbol{y} + \delta\boldsymbol{1}) = \mathbf{P}\boldsymbol{y} + \delta\boldsymbol{1}, \forall \delta \in \mathbb{R}
    &\Leftrightarrow& 
    \mathbf{P}\boldsymbol{1} = \boldsymbol{1}\\
    &\Leftrightarrow& \frac{\boldsymbol{\epsilon} \boldsymbol{v}^T}{\boldsymbol{v}^T \boldsymbol{\epsilon}} \boldsymbol{1} = \boldsymbol{0}
    \label{eq:transl_inv_Constraint}
\end{align}
where $\boldsymbol{1}$ denotes the all one vector. Since $\boldsymbol{\epsilon} \neq 0$ and assuming that $\boldsymbol{v}^T \boldsymbol{\epsilon} \neq 0$,\footnote{Once we have chosen $\boldsymbol{v}$, we will verify that this constraint is satisfied.} 
\eqref{eq:transl_inv_Constraint} implies that  
\begin{equation}
    \boldsymbol{1}^T\boldsymbol{v} = 0 \quad \iff \quad \frac{1}{N}\sum_{i=1}^{N} v_i = 0,
    \label{eq:zeromean_constraint}
\end{equation}
i.e.\ $\boldsymbol{v}$ must be zero-mean.

We now derive $\boldsymbol{v}$. We choose the projection $\mathbf{P}$ such that the approximation is as close as possible to the true codec under the constraint of intensity-shift invariance or equivalently under the constraint that $\boldsymbol{v}$ is centered. In other words,
\begin{equation}
    \boldsymbol{v}^* = \arg \min_{\boldsymbol{v}:\  \boldsymbol{1}^T{\boldsymbol{v}}=0}  \|  \phi(\boldsymbol{y}) - \hat \phi(\boldsymbol{y}) \|^2
    \label{eq:optimPb}
\end{equation}
Let us define the estimation error as $\boldsymbol{r} = \phi(\boldsymbol{y}) - \hat \phi(\boldsymbol{y}) $. From \eqref{eq:def_codec} and \eqref{eq:projector} with $\boldsymbol{u}=\boldsymbol{\epsilon}$, we get
\begin{equation}
    \boldsymbol{r} = (\boldsymbol{\hat{y}} - \boldsymbol{y}) + \boldsymbol{\epsilon} \frac{\boldsymbol{v}^T \boldsymbol{y}}{\boldsymbol{v}^T \boldsymbol{\epsilon}}
    = \boldsymbol{\epsilon} \left( 1 + \frac{\boldsymbol{v}^T \boldsymbol{y}}{\boldsymbol{v}^T \boldsymbol{\epsilon}} \right)
    = \boldsymbol{\epsilon} \frac{\boldsymbol{v}^T \boldsymbol{\hat{y}}}{\boldsymbol{v}^T \boldsymbol{\epsilon}}.
\end{equation}
Therefore, \eqref{eq:optimPb} becomes
\begin{equation}
    \boldsymbol{v}^* = \arg \min_{\boldsymbol{v}:\  \boldsymbol{1}^T{\boldsymbol{v}}=0}  
    \| \boldsymbol{\epsilon} \|^2 \left( \frac{\boldsymbol{v}^T \boldsymbol{\hat{y}}}{\boldsymbol{v}^T \boldsymbol{\epsilon}} \right)^2
\end{equation}
Since $\boldsymbol{\epsilon} \neq 0$, and assuming that $\boldsymbol{v}^T \boldsymbol{\epsilon} \neq 0$,\footnote{Note that this assumption is the same as the one needed to derive \eqref{eq:zeromean_constraint}.} the minimum of the function to be optimized is 0 and can be achieved with $\boldsymbol{v}^*$ that belongs to the intersection of these two hyperplanes:
\begin{equation}
    \boldsymbol{v}^* \in \{\boldsymbol{v}: \boldsymbol{1}^T{\boldsymbol{v}}=0\} \cap \{\boldsymbol{v}: \boldsymbol{v}^T \boldsymbol{\hat{y}}=0\} 
    \label{eq:intersection}
\end{equation}
We propose to set $\boldsymbol{v}^*$ as the centered codec error
\begin{equation}
    \boldsymbol{v}^* = \boldsymbol{\epsilon} - \bar{\boldsymbol{\epsilon}}
\end{equation}
since it lies in the set of optimal solutions defined in \eqref{eq:intersection}. Indeed, $\boldsymbol{\epsilon} - \bar{\boldsymbol{\epsilon}}$ is centered, and is orthogonal to the codec output, as it was observed in Section~\ref{sec:codecProp}. Moreover, this choice of $\boldsymbol{v}$ satisfies $\boldsymbol{v}^T \boldsymbol{\epsilon} \neq 0$ since the mean $\bar{\boldsymbol{\epsilon}}$ is close to 0, and the vectors $\boldsymbol{v}^* = \boldsymbol{\epsilon} - \bar{\boldsymbol{\epsilon}}$ and $\boldsymbol{\epsilon}$ are almost colinear.

To summarize, we proposed a first-order codec model, that satisfies the properties of the true codec presented in Section~\ref{sec:codecProp}. The model is a linear projector, whose equation is 
\begin{equation}
    \hat \phi(\boldsymbol{{y}}) 
    = \left(\mathbf{I} - \frac{\boldsymbol{\epsilon}\,(\boldsymbol{\epsilon} - \bar{\boldsymbol{\epsilon}})^T}{N\sigma^2_{\boldsymbol{\epsilon}}}\right)\boldsymbol{y}.
    \label{eq:model}
\end{equation}

\subsection{Gradient of the codec model $\hat \phi$}
We now differentiate the proposed model $\hat \phi$ \eqref{eq:model} with respect to $\boldsymbol{y}$, treating $\boldsymbol{\epsilon}$ 
as independent of $\boldsymbol{y}$. This yields
\begin{equation}
    \mathbf{J}_{\text{Proj}}(\boldsymbol{y}) = \mathbf{I} - 
    \frac{\boldsymbol{\epsilon}\,(\boldsymbol{\epsilon} - \bar{\boldsymbol{\epsilon}})^T}{N\sigma^2_{\boldsymbol{\epsilon}}},
    \label{Jproj}
\end{equation}
%
which is equal to $\mathbf{J}_{\text{SCALED}}$ in~(\ref{eq:sgcodec}b). 
This establishes that SCALED~\cite{pesnel2025scaled} is the surrogate gradient of the MSE-optimal projector under the observed constraints of true video codecs, such as intensity-shift invariance, non-centered compression error, and orthogonality between the centered compression error and the codec output.


\subsection{Comparison with learned proxies}

From this projection perspective, we can derive a set of properties explaining the superiority of the surrogate gradient in (\ref{Jproj}) over learned proxies.

\textbf{Stability:} By modeling the codec Jacobian $\mathbf{J}_{\text{Proj}}$ 
as a geometric projection, our method inherits the fundamental property of 
\textit{idempotence} ($\mathbf{P}^2=\mathbf{P}$). This imposes a deterministic 
binary structure on the Jacobian's spectrum, where eigenvalues are strictly 
$\lambda \in \{0, 1\}$, fixing the spectral radius at 
$\rho(\mathbf{J}_{\text{Proj}}) = 1$. This bounded spectrum has a concrete 
consequence on the gradient flow: components along the directions associated 
with $\lambda = 1$ are preserved unaltered, while those associated with 
$\lambda = 0$, i.e. corresponding to gradient directions irrelevant to the 
codec output, are filtered out. This intrinsic boundedness prevents the 
gradient explosion issues commonly observed with unconstrained surrogate 
approximations of non-differentiable codec operations.

\textbf{Zero-shot adaptability:} The projection operator is parameter-free, depending solely on the statistics of $\boldsymbol{\epsilon}$. It adapts instantly to any QP or codec configuration change: as the error magnitude or distribution shifts, the projection hyperplane adjusts automatically.

\begin{table}[h]
    \caption{Intensity-shift invariance of two codecs (\textit{medium} preset). Values report $\mathbb{E}[\phi(\boldsymbol{x}+\delta) - \phi(\boldsymbol{x}) - \delta]$ averaged over 30 sequences from~\cite{Google2024}; $0$ indicates invariance.}
   \label{tab:translation_invariance}
   \centering
    \renewcommand{\arraystretch}{1.2}
    \setlength{\tabcolsep}{5pt}
    \resizebox{\linewidth}{!}{
        \begin{tabular}{c c cccccc}
        \toprule
        \multirow{2}{*}{\textbf{QP}} 
        & \multirow{2}{*}{\textbf{Codec}}
        & \multicolumn{6}{c}{\textbf{Translation offset} $\delta$} \\
        \cmidrule(lr){3-8}
        & & $-5$ & $-3$ & $-1$ & $+1$ & $+3$ & $+5$ \\
        \midrule
        \multirow{2}{*}{32} 
        & x264   & 0.015  & 0.002  & -0.002 & -0.012 & -0.007 & -0.027 \\
        & VVenC  & 0.007  & 0.017  & -0.005 & -0.005 & -0.011 & -0.018 \\
        \midrule
        \end{tabular}
    }
\end{table}

\textbf{Fidelity:} Traditional codec proxies often fail to capture the behavior of authentic codecs due to modeling complexity. To ensure differentiability, many solutions rely on simplifying assumptions (e.g., fixed $8 \times 8$ partitioning, simplified intra-modes, or optical flow instead of block matching~\cite{Google2024}). These discrepancies lead to a domain gap, causing suboptimal performance when transferring from training to inference~\cite{pesnel2025scaled}. In contrast, our approach relies on the true codec execution in the forward pass and a constrained MSE-optimal projection (under the proposed projection formulation) in the backward pass.

\section{Application to Neural Wrapper Design}

Although the SCALED surrogate gradient has been empirically validated for learned downscalers \cite{pesnel2025scaled}, its extension to full neural wrappers, which jointly train a pre- and a post-processor, has not yet been established. We address this gap in the next section.

\subsection{Neural wrapper architecture}

Although compatible with various backbones, we employ a U-Net~\cite{U-Net} for both pre- and post-processing tasks. To foster convergence, the network operates strictly in the residual domain. Following the design in~\cite{Google2024}, the encoder comprises a filter sequence of $(32, 64, 128, 256)$, leading to a 512-channel bottleneck, operating in YUV444 format. Resolution scaling is handled via deterministic filters, bicubic downscaling and Lanczos upscaling, allowing both neural models to operate effectively at full input resolution.

\begin{table*}[t]
    \centering
    \caption{Comparison of BD-Rate performance using x264 and VVenC. 
    Reference is codec if $s=1$, else $Lanczos\downarrow Bicubic\uparrow$.}
    \label{tab:results}
    
    \renewcommand{\arraystretch}{1.2}
    \setlength{\tabcolsep}{4pt}
    \resizebox{\linewidth}{!}{
        \begin{tabular}{c l cccc cccc}
        \toprule
        \multirow{2}{*}{\textbf{Ratio}} 
        & \multirow{2}{*}{\textbf{Method (Forward / Backward)}}
        & \multicolumn{4}{c}{\textbf{x264 (H.264/AVC) \cite{x264}}}
        & \multicolumn{4}{c}{\textbf{VVenC (H.266/VVC) \cite{vvenc}}} \\
        \cmidrule(lr){3-6} \cmidrule(lr){7-10}
        & & \textbf{PSNR} & \textbf{SSIM} & \textbf{VMAF} & \textbf{VMAF-NEG}
          & \textbf{PSNR} & \textbf{SSIM} & \textbf{VMAF} & \textbf{VMAF-NEG} \\
        \midrule

        \multirow{3}{*}{\textbf{$\times 1$}}
        & Proxy \cite{Google2024} / Proxy \cite{Google2024}      & -0.56\% & -0.38\% & 5.21\%  & 3.18\%  & 1.66\%  & 5.02\%  & -1.18\%  & -1.33\% \\
        & Proxy \cite{Google2024} / Proxy+Proj. quant. (Ours)          & -0.21\% & -11.96\% & 2.53\%  & 1.30\% & -0.15\%  & -0.06\%  & -0.12\%  & -0.11\% \\
        & True codec / Surrogate (Ours)                          & \textbf{-7.70\%} & \textbf{-12.45\%} & \textbf{-4.25\%} & \textbf{-4.01\%} & \textbf{-1.34\%} & \textbf{-2.14\%} & \textbf{-1.91\%} & \textbf{-1.65\%} \\
        \midrule\midrule

        \multirow{3}{*}{\textbf{$\times 1/2$}}
        & Proxy \cite{Google2024} / Proxy \cite{Google2024}      & -5.33\% & -7.82\% & 4.81\%  & 3.61\%  & -2.11\%  & 2.11\%   & -4.16\%  & -4.66\% \\
        & Proxy \cite{Google2024} / Proxy+Proj. quant. (Ours)          & -8.30\% & -8.39\% & -0.87\% & -1.85\% & -2.50\%  & 0.39\%   & -0.82\%  & -1.87\% \\
        & True codec / Surrogate (Ours)                          & \textbf{-15.66\%} & \textbf{-11.78\%} & \textbf{-11.90\%} & \textbf{-11.68\%} & \textbf{-8.17\%} & \textbf{-4.76\%} & \textbf{-5.81\%} & \textbf{-5.72\%} \\
        \midrule\midrule

        \multirow{3}{*}{\textbf{$\times 1/4$}}
        & Proxy \cite{Google2024} / Proxy \cite{Google2024}      & -10.62\% & -13.14\% & -10.35\% & -10.91\% & -8.40\%  & -8.57\%  & -15.21\% & -15.39\% \\
        & Proxy \cite{Google2024} / Proxy+Proj. quant. (Ours)           & -15.99\% & -18.97\% & -8.40\%  & -9.28\%  & -9.70\%  & -8.16\%  & -9.99\%  & -10.95\% \\
        & True codec / Surrogate (Ours)                          & \textbf{-23.59\%} & \textbf{-21.01\%} & \textbf{-36.17\%} & \textbf{-35.97\%} & \textbf{-20.07\%} & \textbf{-13.47\%} & \textbf{-30.56\%} & \textbf{-30.55\%} \\

        \bottomrule
        \end{tabular}
    }
\end{table*}

\begin{figure}
    \centering
    \includegraphics[width=0.85\linewidth]{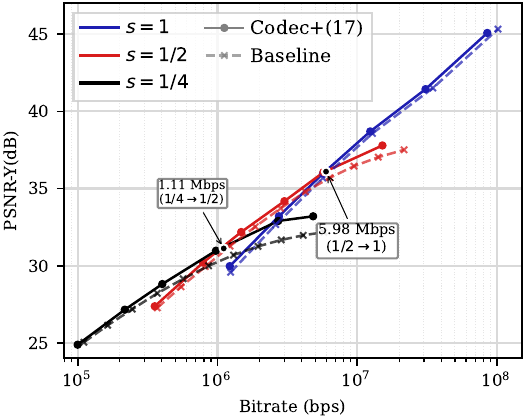}
    \vspace{-0.3cm}
    \caption{Convex hull comparison of average RD curves for various encoding scales (x264, \textit{medium} preset)}
    \label{fig:side_by_side}
    \vspace{-0.5cm}
\end{figure}

\subsection{Learning strategies for comparison}
\subsubsection{Standard codec proxy}
To demonstrate the effectiveness of our approach, we primarily rely on the codec proxy $\hat{\phi}$ from the sandwiched compression framework \cite{Google2024}, which is based on the hybrid coding paradigm. Quantization is modeled with a Straight-Through Estimator (STE) \cite{STE}. Motion estimation is performed offline using an optical flow model (U-Flow \cite{UFlow}). Finally, the rate is estimated using a rate proxy $\hat{\psi}$, designed as a differentiable approximation of the $\ell_0$ norm of DCT coefficients. Consequently, we jointly optimize $f$ and $g$ by minimizing the following objective:

\begin{equation}
\begin{split}
    \boldsymbol{\theta}_{f}^{*}, \boldsymbol{\theta}_{g}^{*} &= 
    \arg\min_{\boldsymbol{\theta}_f, \boldsymbol{\theta}_g} \left\| \boldsymbol{x} - g \left(\hat{\phi}(f(\boldsymbol{x};\boldsymbol{\theta}_f));\boldsymbol{\theta_g} \right) \right\|^2_2 \\
    &\quad ~~~~~~~~~~~~~~~~~~~~~~~~~~~~~ + \lambda\hat{\psi}(f(\boldsymbol{x}; \boldsymbol{\theta}_f))
\end{split}
\label{eq:joint_optimization}
\end{equation}
where $\lambda$ is a Lagrange multiplier controlling the rate-distortion tradeoff.

\subsubsection{Codec proxy coupled with projection-based surrogate for quantization}

A critical drawback of the previous method is the STE-based quantization, which causes training divergence at low bit rates. To mitigate this limitation, we substitute the STE quantization modeling used in \cite{Google2024} with our projection-based surrogate gradient of \eqref{Jproj} since it is applicable to any non-differentiable block such as quantization. Consequently, the new quantization formulation adopted within the codec proxy is now:

\begin{equation}
    \begin{cases}
    \text{Forward}: & \bar{\boldsymbol{x}} = \boldsymbol{x} +\boldsymbol{Q}_e \\
    \text{Backward}: & \dfrac{\partial \bar{\boldsymbol{x}}}{\partial \boldsymbol{x}} = \mathbf{I} - \dfrac{\boldsymbol{Q}_e(\boldsymbol{Q}_e-\bar{\boldsymbol{Q}_e})^T}{N\sigma^2_{\boldsymbol{Q}_e}}
    \end{cases}
    \label{eq:scaled_ste}
\end{equation}
where $\boldsymbol{Q}_e = \left\lfloor \frac{\boldsymbol{x}}{Q_{\text{step}}} \right\rceil \cdot Q_{\text{step}} - \boldsymbol{x}$ is the quantization noise.

Here, $\sigma_{\boldsymbol{Q}_e}$ represents the standard deviation of the quantization error estimated on the frame output by the proxy. This stabilizes the gradient flow, thus preventing divergence issues commonly observed with the STE at low bit rates.

\subsubsection{Neural wrappers with projection-based surrogate gradient}

We fully replace here any distortion-based codec proxy with the projection-based codec surrogate gradient learning strategy presented in Section~\ref{sec:demo}, using the true codec error at training. Specifically, we employ a rate-distortion loss in the subsequent experiments, minimizing the following cost function:

\begin{equation}
\begin{split}
    \boldsymbol{\theta}_{f}^{*}, \boldsymbol{\theta}_{g}^{*} &= \arg\min_{\boldsymbol{\theta}_f, \boldsymbol{\theta}_g} \left\| \boldsymbol{x} - g\left(\boldsymbol{\hat{y}} ; \boldsymbol{\theta}_g\right) \right\|^2_2 + \lambda\hat{\psi}(f(\boldsymbol{x}; \boldsymbol{\theta}_f))
\end{split}
\label{eq:scaled_optimization}
\end{equation}
where $\hat{\psi}$ denotes the same standard rate proxy as used in ~\cite{Google2024} and previous sections.
\section{Experiments}
\subsection{Experimental Setup}
\textbf{Training details:} We train independent models for each scale ratio $s\in\{1, 1/2, 1/4\}$ and target QP $\in \{17, 23, \dots, 47\}$ using the Google Sandwiched Compression dataset~\cite{Google2024} (YUV 4:4:4 sequences, $256\times256$, 10 frames). Input chroma is subsampled and upsampled to simulate YUV 4:2:0. Training runs for 500 epochs using Adam~\cite{Adam} ($\eta=10^{-4}$) with a weighted MSE loss (4:1:1).

\textbf{Evaluation:}
We evaluate performance on 24 Xiph and 7 UVG~\cite{UVG} sequences at 1080p resolution. BD-Rate (PSNR, SSIM, VMAF, VMAF-NEG) is reported for different resolution-scale encoding settings: (i) original-resolution coding ($s=1$) with neural pre-/post-processing, compared to standalone coding; and (ii) sub-scale coding ($s \in \{1/2, 1/4\}$) with neural pre-/post-processing, compared to standard coding with Lanczos and bicubic down/upscaling baselines. We evaluate neural wrapping performance across two codecs spanning a wide range of coding efficiency: H.264/AVC~\cite{x264} implemented via x264, and H.266/VVC~\cite{vvenc} implemented via VVenC.

\subsection{Performance assessment}

Table~\ref{tab:results} reports BD-Rate results across all methods, scale ratios, and codecs. Several consistent trends emerge.

\textit{Gains generalize across diverse settings.} Improvements are observed for both x264 and VVenC, and across all scale ratios ($\times 1$, $\times 1/2$, $\times 1/4$), demonstrating the robustness and versatility of the proposed approach.

\textit{Gains extend beyond the training metric.} Although the filters are optimized under an MSE criterion, consistent improvements are also observed for perceptual metrics such as SSIM, VMAF, and VMAF-NEG, with BD-Rate reductions reaching up to $-36.17\%$ (VMAF, $\times 1/4$, x264).

\textit{Our surrogate consistently outperforms proxy-based methods}, yielding systematically higher gains than the standard codec proxy~\cite{Google2024}, both when the proxy is used alone and when it is combined with our projection-based quantization surrogate. The latter variant, where only the STE quantization block is replaced, already improves over the baseline proxy across all configurations, demonstrating that the proposed surrogate generalizes beyond codecs to arbitrary non-differentiable operations. Two complementary factors explain the superiority of the full codec surrogate gradient: the use of the true decoded signal in the forward pass and the use of true codec noise realizations in the backward pass. Together with the linear projection model derived in Section~\ref{sec:demo}, they account for the strong numerical performance observed here.

\textit{Gains decrease with codec efficiency.} The contribution of neural filters is, as expected, lower for VVenC than for x264, reflecting the superior coding efficiency of H.266/VVC. 

\textit{Neural wrapping at $\times 1$.} At full resolution, gains are significant for x264 across all metrics, while remaining more limited for VVenC, reflecting the near-optimal MSE performance of H.266/VVC. This aligns with recent findings suggesting that perceptual-oriented losses may be better suited for recent codecs~\cite{khan2025perceptual}.

We plot the average RD curves in Figure~\ref{fig:side_by_side}, using x264. Our method yields superior performance at every operating point, and the observed switching points validate the use of downscaling for bit rates under $5.98$~Mbps, consistent with bit rate ladder constructions.

\vspace{-0.15cm}

\section{Conclusion}

In this work, we presented a geometric interpretation of surrogate gradients 
for the end-to-end optimization of neural video codec wrappers. By modeling 
the codec as a linear projection, we derived analytically the SCALED 
surrogate gradient as the projection vector that is both MSE-optimal 
and consistent with the intensity-shift invariance empirically observed in 
standard video codecs. 
Moreover, we demonstrated
that this surrogate gradient extends successfully to the 
challenging task of full neural wrapper training with joint pre- and 
post-processing, achieving BD-Rate (PSNR) gains of up to $-23.59\%$ 
on x264 and $-20.07\%$ on VVenC across multiple scale ratios. Future work 
may investigate the extension of this geometric framework to perceptual 
loss functions, motivated by the residual gap observed at full resolution 
with state-of-the-art codecs.

\vspace{-0cm}

\bibliographystyle{IEEEtran}
\bibliography{refs} 

\end{document}